\pdfoutput=1

\documentclass[11pt]{article}

\usepackage[preprint]{acl}

\usepackage{times}
\usepackage{latexsym}
\usepackage{amsmath}

\usepackage[T1]{fontenc}

\usepackage[utf8]{inputenc}

\usepackage{microtype}

\usepackage{inconsolata}

\usepackage{graphicx}

\usepackage{tabularx}
\usepackage{tablefootnote}
\usepackage{subcaption}
\title{Human-LLM Coevolution: Evidence from Academic Writing}

\author{%
  Mingmeng Geng$^{1}$ \quad Roberto Trotta$^{1,2}$\\
  $^1$International School for Advanced Studies (SISSA) \quad $^2$Imperial College London \\
  \texttt{mgeng@sissa.it}
}

\begin{document}
\maketitle
\begin{abstract}
With a statistical analysis of arXiv paper abstracts, we report a marked drop in the frequency of several words previously identified as overused by ChatGPT, such as ``delve'', starting soon after they were pointed out in early 2024. The frequency of certain other words favored by ChatGPT, such as ``significant'', has instead kept increasing. These phenomena suggest that some authors of academic papers have adapted their use of large language models (LLMs), for example, by selecting outputs or applying modifications to the LLM outputs. Such coevolution and cooperation of humans and LLMs thus introduce additional challenges to the detection of machine-generated text in real-world scenarios. Estimating the impact of LLMs on academic writing by examining word frequency remains feasible, and more attention should be paid to words that were already frequently employed, including those that have decreased in frequency due to LLMs' disfavor.
\end{abstract}

\section{Introduction}

After the launch of ChatGPT at the end of 2022, large language models (LLMs) began to be widely used and are now transforming many aspects of our work and life, with academic writing being one of them. The coevolution of AI and humans has also been recognized by researchers~\citep{pedreschi2024human}.

For example, empirical studies from April 2024 observed that the frequency of certain words used in academic papers published in 2023 had changed and confirmed a strong correlation between these changes and the use of LLMs~\citep{liang2024mapping,geng2024chatgpt}. Survey results also show that many researchers are utilizing LLMs in their work~\citep{liao2024llms}. 

The detection of machine-generated text (MGT) has also attracted a lot of attention~\citep{tang2024science,chowdhury2024genai,wang2025genai}, but the performance of detectors has also been questioned early on~\citep{sadasivan2023can,weber2023testing,ghosal2023towards}. Recent studies continue to show that some methods are not sufficiently robust~\citep{zhang2024detection,wang2024stumbling,creo2025silverspeak}. The effectiveness of MGT detectors is also related to the model of LLMs and the type of text~\citep{liu2024generalization}, and their accuracy may also be exaggerated~\citep{doughman2024exploring}. The situations likely to arise in reality are more complicated and are not limited to a binary classification framework~\cite{zhang2024llm}. Thus, examining and analyzing the ongoing evolution of word usage remains a useful and meaningful task.

\begin{figure}[t]
\includegraphics[width=\columnwidth]{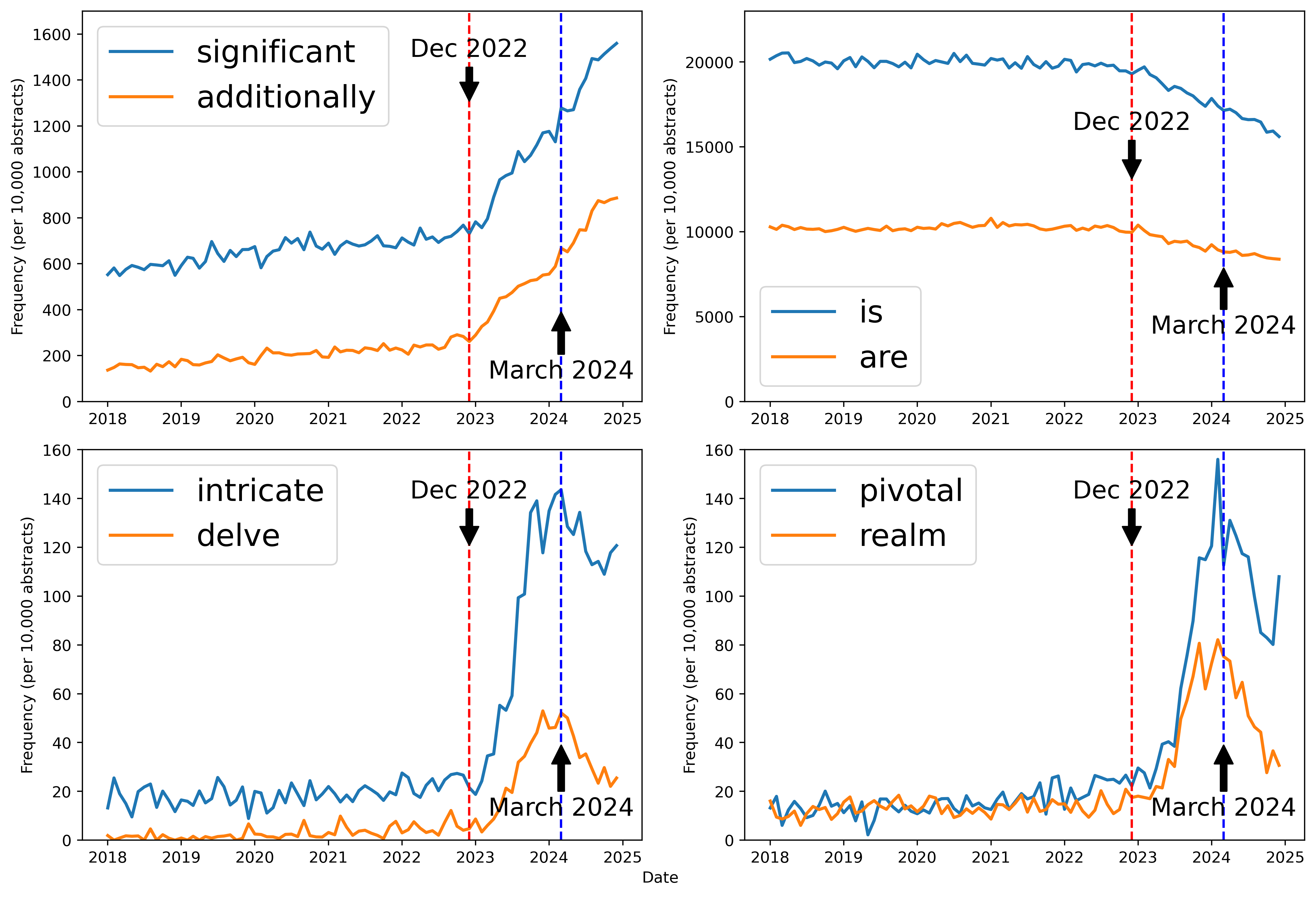}
  \caption{The frequency evolution of some words in arXiv abstracts (they were singled out around April 2024 as either favored or disfavored by ChatGPT).}
  \vspace{-2em}
  \label{wf_example}
\end{figure}

\begin{figure*}[t]
    \centering
    \begin{subfigure}[b]{0.32\textwidth}
        \centering
        \includegraphics[width=\textwidth]{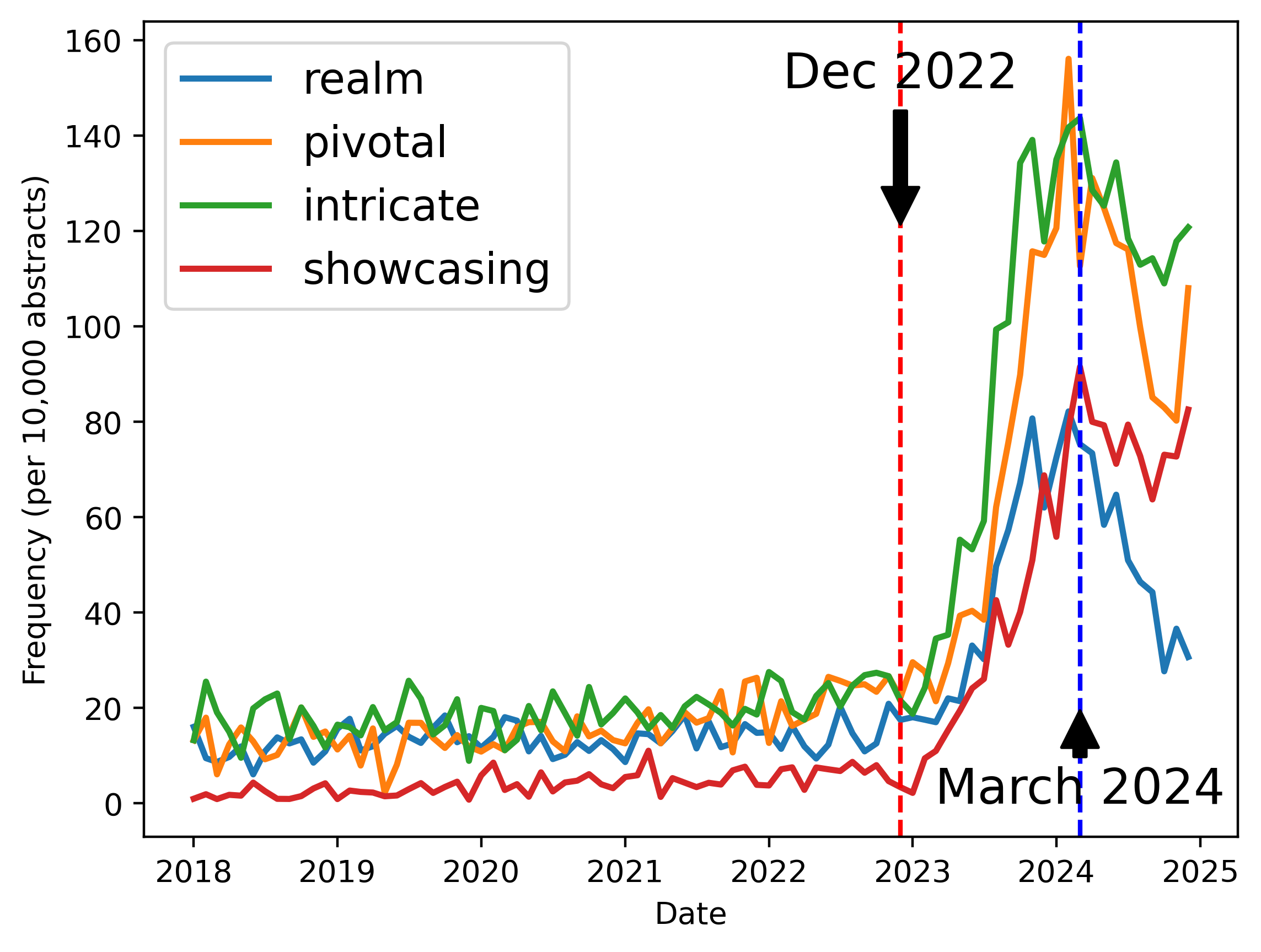}
        \caption{}
        \label{wf_realm}
    \end{subfigure}
    \hfill
    \begin{subfigure}[b]{0.32\textwidth}
        \centering
        \includegraphics[width=\textwidth]{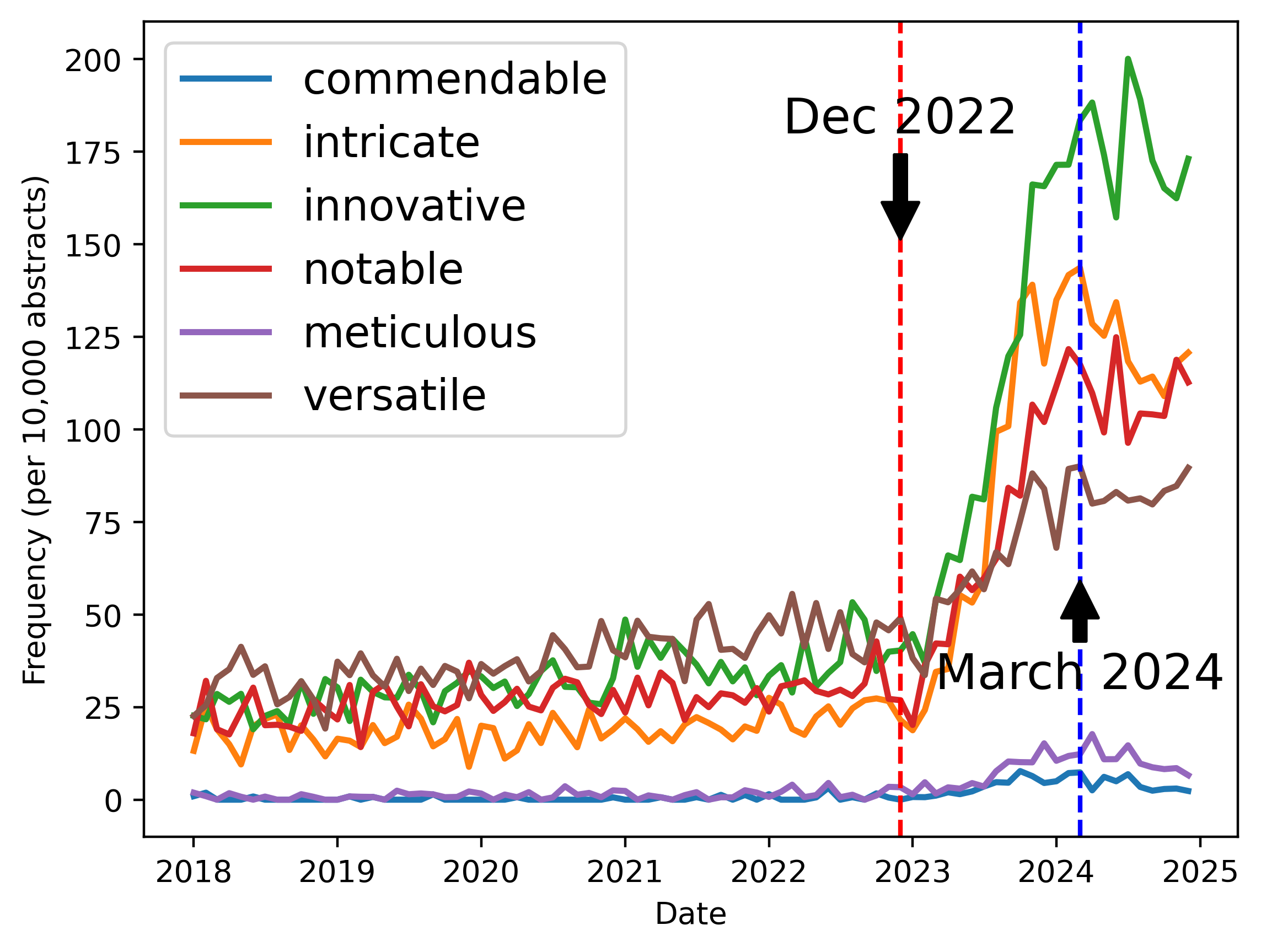}
        \caption{}
        \label{wf_commenable}
    \end{subfigure}
    \hfill
    \begin{subfigure}[b]{0.32\textwidth}
        \centering
        \includegraphics[width=\textwidth]{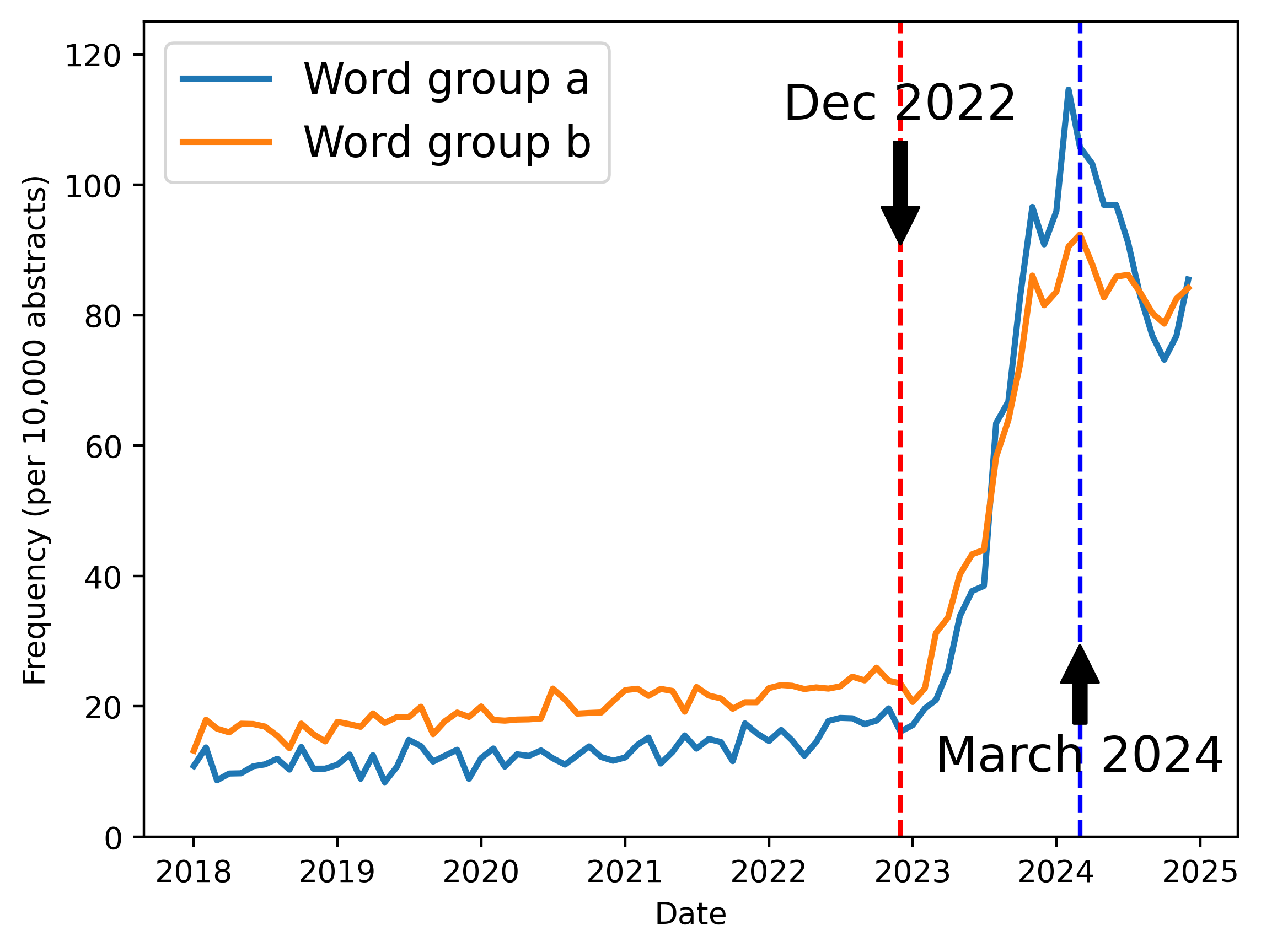}
        \caption{}
        \label{wf_group}
    \end{subfigure}
    \hfill
    \begin{subfigure}[b]{0.32\textwidth}
        \centering
        \includegraphics[width=\textwidth]{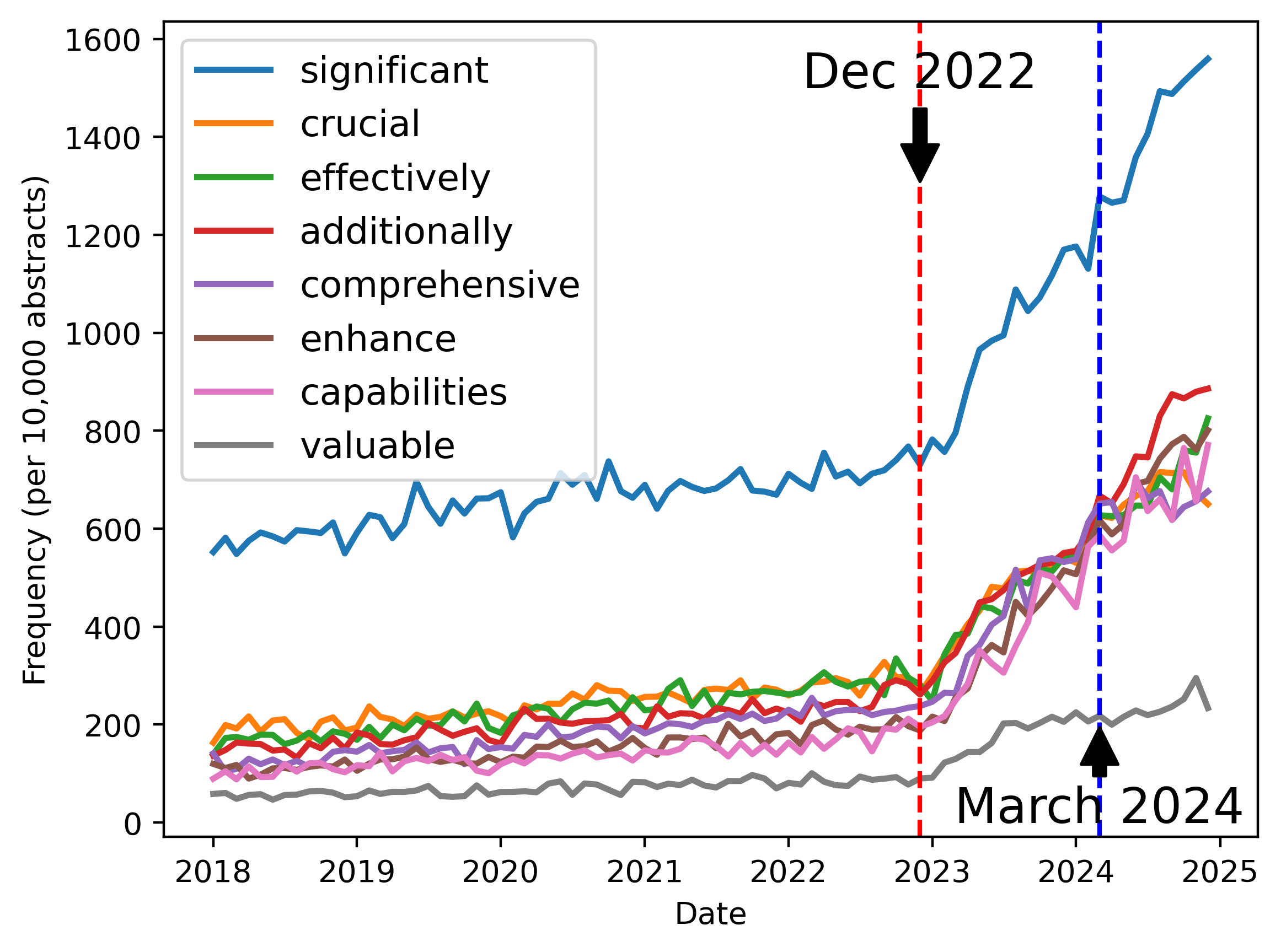}
        \caption{}
        \vspace{-0.5em}
        \label{wf_significant}
    \end{subfigure}
    \hfill
    \begin{subfigure}[b]{0.32\textwidth}
        \centering
        \includegraphics[width=\textwidth]{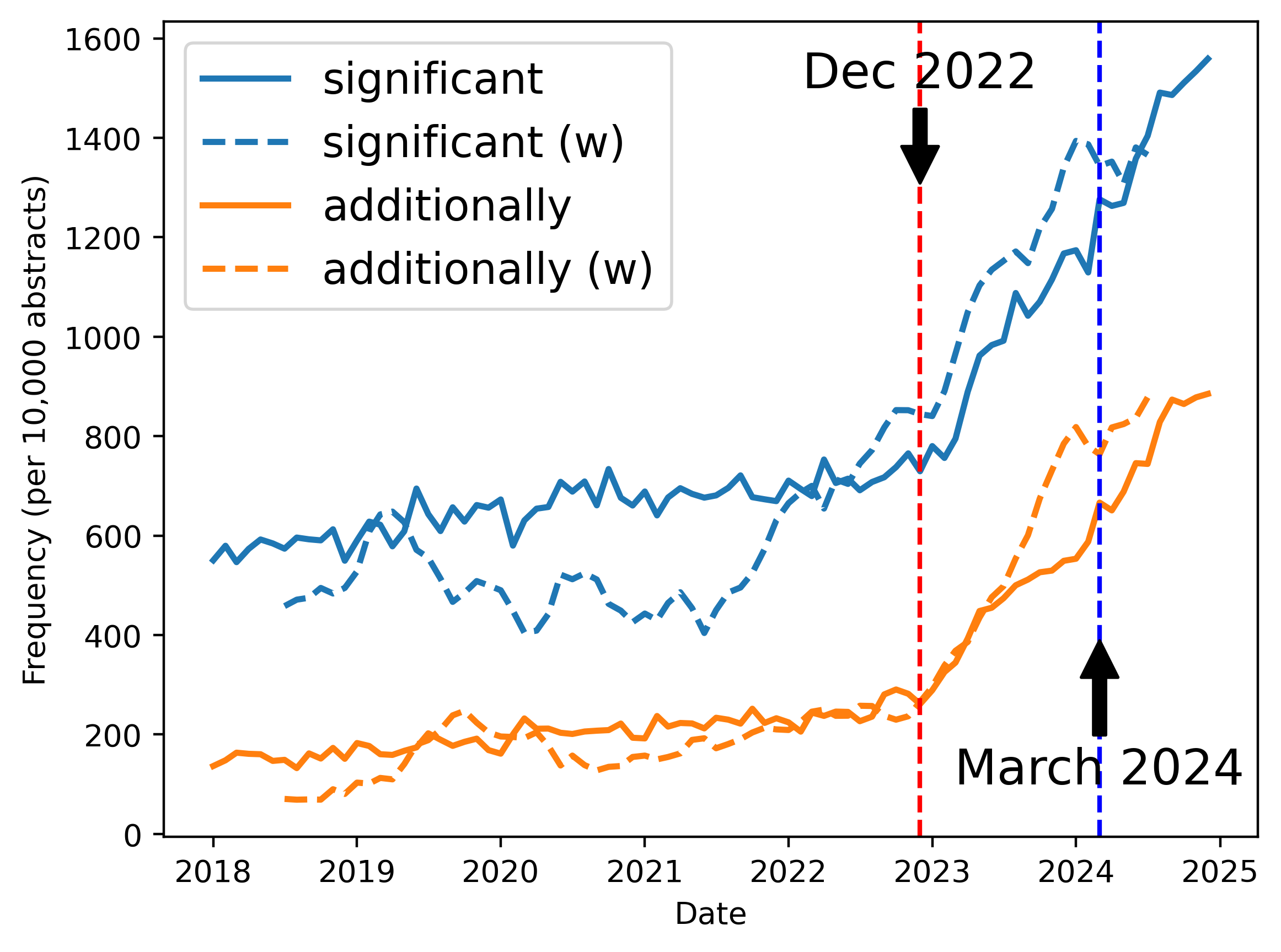}
        \caption{}
        \vspace{-0.5em}
        \label{wf_example_w1}
    \end{subfigure}
    \hfill
    \begin{subfigure}[b]{0.32\textwidth}
        \centering
        \includegraphics[width=\textwidth]{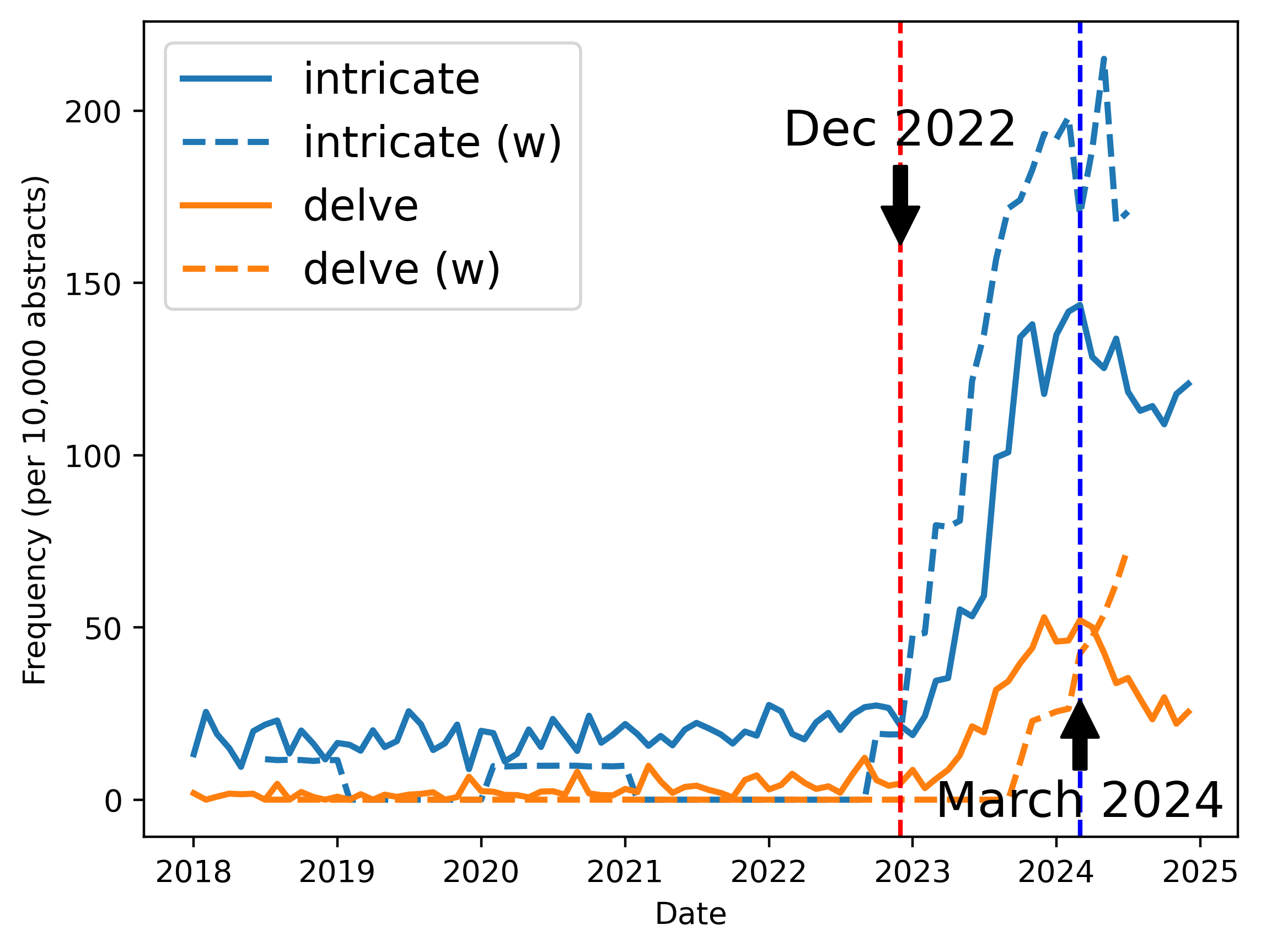}
        \caption{}
        \vspace{-0.5em}
        \label{wf_example_w2}
    \end{subfigure}
    \caption{Frequency of words in arXiv abstracts previously identified as indicative of LLM usage. All word frequencies are normalized based on 10,000 abstracts. Word groups a and b correspond to the average frequencies of the words in \ref{wf_realm} and \ref{wf_commenable}. The data for withdrawn papers represents a 12-month rolling average, labeled by ``w''.}
    \vspace{-1em}
  \label{wf_ex_more}
\end{figure*}

Figure~\ref{wf_example} illustrates the evolution in the frequency usage of some of the words that were singled out as favored or disfavored by ChatGPT around April 2024. The frequency of ``significant'' and ``additionally'' continues to grow, while that of ``is'' and ``are'' continues its declining trend, as noted by~\citet{geng2024chatgpt}. Meanwhile, the frequency of some other words (e.g., ``intricate'' and ``delve'') associated with LLMs begins to decrease after March and April 2024, which corresponds to the time when researchers identified these words in AI conference peer reviews~\citep{liang2024monitoring} and academic papers~\citep{liang2024mapping}.

Changes in the words used in academic writing, as discussed above, serve as an excellent example of AI and human coevolution. Researchers are constantly proposing new detection techniques, but the language and expressions of LLM users are also likely evolving due to their use of LLMs~\citep{geng2024impact}. Given the lack of a precise definition, LLM-generated text might be undetectable in certain individual instances. Therefore, statistically measuring the impact of LLMs over a large corpus of texts is a more practical option.

This paper focuses on the following key points:
\begin{itemize}
\item The different fates of word frequencies after changes have been pointed out and scribed to LLMs usage.
    \item The challenges of MGT detectors.
    \item The long-term impact of LLMs in academic writing.
\end{itemize}

\section{Data}
\paragraph{arXiv paper metadata} Metadata of arXiv papers updated weekly on Kaggle\footnote{\url{https://www.kaggle.com/datasets/Cornell-University/arxiv/data}}. Our paper used version 214 of this dataset. Between January 2018 and December 2024, February 2018 and October 2024 recorded the lowest and highest numbers of papers, at 10,593 and 24,226, respectively. During this period, the total number of papers is 1,294,653. 

\paragraph{Withdrawn arXiv papers data} WithdrarXiv dataset~\citep{rao2024withdrarxiv}, containing over 14,000 arXiv withdrawn papers up to September 2024.

\section{Word Frequency Analysis}
\label{wf_section}

The analysis presented in Figure~\ref{wf_ex_more} is based on the abstracts of all arXiv papers submitted between 2018 and 2024. The frequency of words is calculated on a monthly basis and normalized per 10,000 abstracts. 

\begin{figure*}[t]
    \centering
    \begin{subfigure}[b]{0.43\textwidth}
        \centering
        \includegraphics[width=\textwidth]{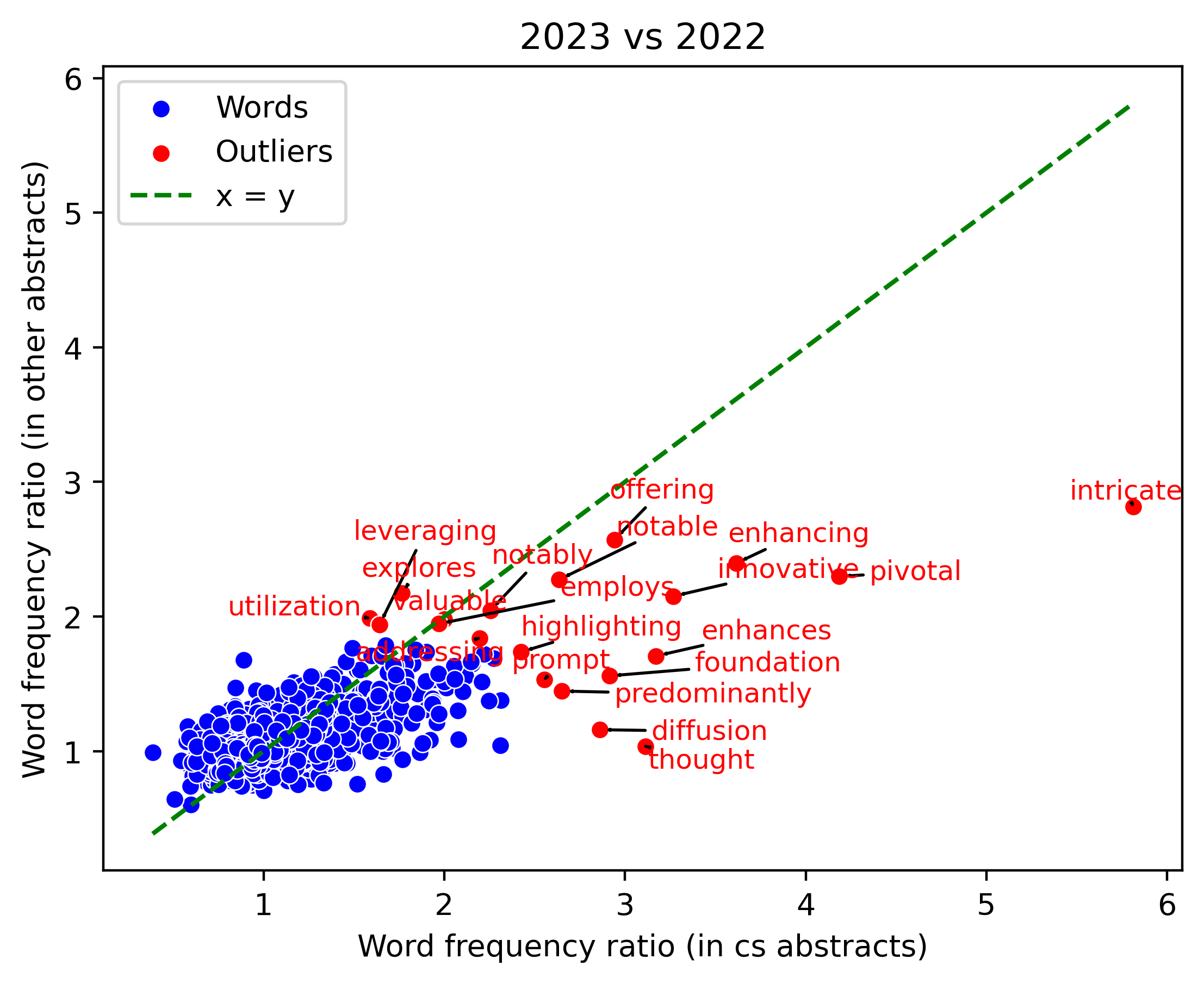}
        \caption{}
        \vspace{-0.5em}
        \label{wf_ratio_23_22}
    \end{subfigure}
    \hfill
    \centering
    \begin{subfigure}[b]{0.43\textwidth}
        \centering
        \includegraphics[width=\textwidth]{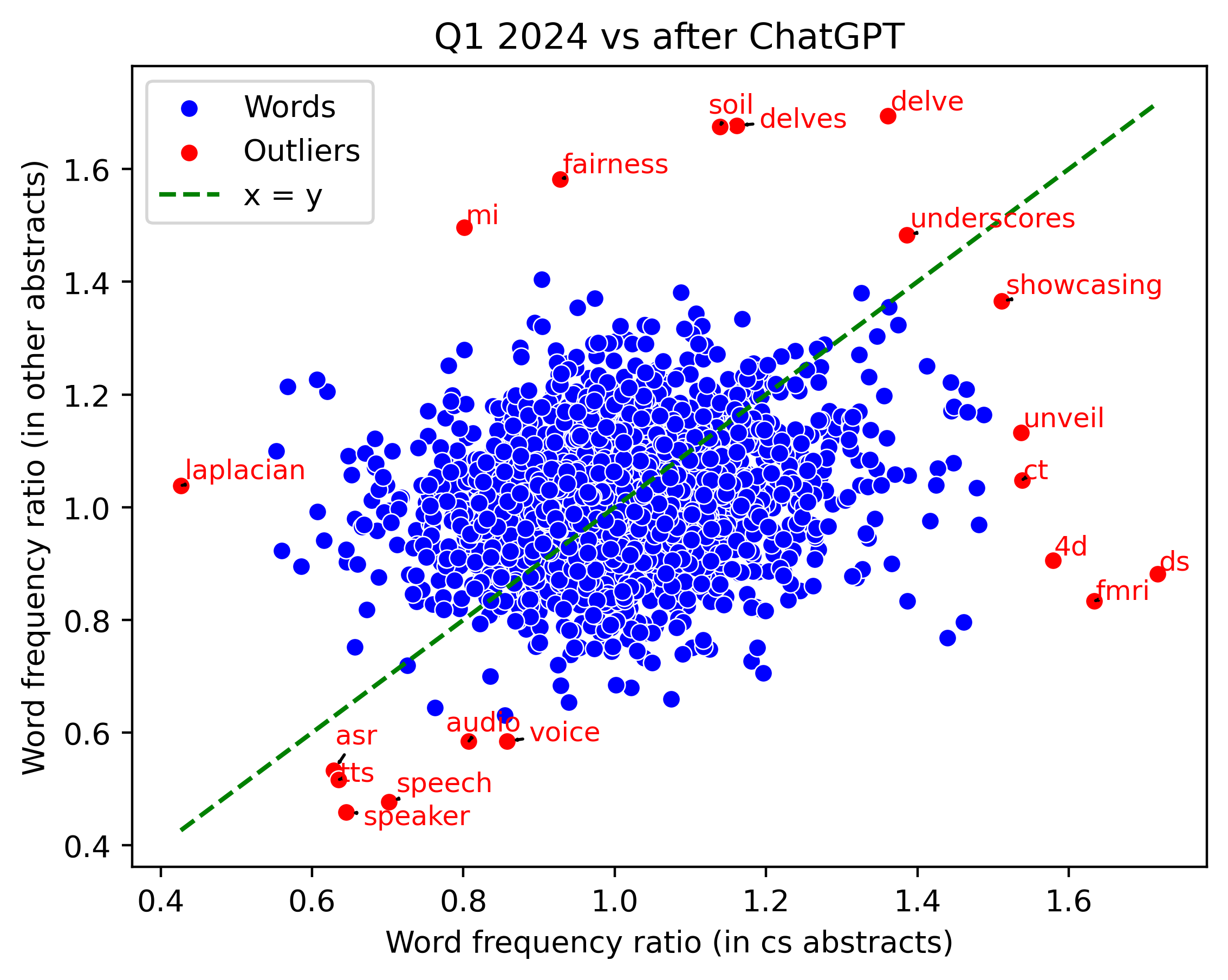}
        \caption{}
        \vspace{-0.5em}
        \label{wf_ratio_q1_2024}
    \end{subfigure}
    \caption{Comparing the ratio of word frequency between Computer Science abstracts and other disciplines. Only words that appear at least 20 times on average per 10,000 abstracts are plotted.}
    \vspace{-1em}
  \label{wf_cs_other}
\end{figure*}

Figure~\ref{wf_realm} shows the frequency of the 4 words highlighted by~\citet{liang2024mapping} and Figure~\ref{wf_commenable} presents the frequency of the 6 words emphasized by~\citet{liang2024monitoring}. The former paper analyzes academic papers, while the latter focuses on AI conference peer reviews, and the average frequency of these words is shown in Figure~\ref{wf_group}. The trend is clear: starting from April 2024, the frequency of these well-known LLM-style words began to decrease. Some other words show patterns of consistent growth or a rise followed by a decline, as illustrated in Figure~\ref{wf_advancements} of the Appendix.

A study published in December 2024 also observed a decline in the use of certain words, such as ``delve'', in some selected arXiv papers~\citep{leiter2024nllg}. While they suggested that this was likely due to the release of GPT-4o in May 2024, we suggest that the main reason is that LLMs may have given these words a bad reputation. Many researchers noticed such kind of words in March and April and quickly changed their arXiv abstracts. If new LLMs were the cause, the drop in word frequency would have been delayed.

In addition, the frequency of words like ``significant'', specifically pointed out by~\citet{geng2024chatgpt}, continues to grow. This may be because these terms are relatively common and frequently used, their presence alone would not easily lead one to suspect the text as the product of LLMs. Besides, as presented in Table~\ref{paper_compare}, this article has attracted less attention than the former, for example, in terms of Google Scholar citation counts. Therefore, fewer researchers should have noticed the relationship between these words and LLMs.

We compared the results with the abstracts of the withdrawn papers, as illustrated in Figures~\ref{wf_example_w1} and \ref{wf_example_w2}. Given the small number of withdrawn papers, the 12-month rolling averages of their word frequency are used in the graphs. The frequency of some words, such as ``intricate'', is higher in the withdrawn papers, but the difference is not very large, as is also the case in Figures \ref{wf_realm_w}, \ref{wf_crucial_w} and \ref{wf_is_w} of the Appendix.

To better compare the changes in word frequency, we define $R_{ij}(T_1,T_2)$ (the ratio of word $i$ in the abstracts of category $j$ between periods $T_1$ and $T_2$) as follows: $ R_{ij}(T_1,T_2) = \frac{f_{ij}(T_1)}{f_{ij}(T_2)} $, where $f_{ij}(T)$ is the frequency of word $i$ in the abstracts of category $j$ during the time period $T$.

We also categorized the abstracts into two groups based on the first category of the papers: computer science (\textit{cs}) and others.  Figure~\ref{wf_ratio_23_22} represents the ratio $R$ between 2023 and 2022, where some words, like ``diffusion'', are related to the research topics, but some other words have also become much more common in different fields. The ratio $R$ in Figure~\ref{wf_ratio_q1_2024} is calculated using the word frequency in the first quarter of 2024 divided by the word frequency from January 2023 to December 2024. Some words like ``delve'' and ``showcasing'' actually reached their peak usage from January to March 2024, and such words are very few. Figure~\ref{cs_o} provides more detailed examples. Words that appear more often in \textit{cs} paper abstracts have also clearly increased in other disciplines. 

More researchers have now noticed issues with word usage and diversity in LLM-generated content~\citep{kobak2024delving,reviriego2024playing,guo2024benchmarking}. Based on the above results, people are likely still using LLMs, but they may avoid some words that are typical of LLM output. Therefore, detecting LLM-generated content in real-world scenarios may become more difficult.

\section{Challenges in Machine-Generated Text Detection}

The first 1000 arXiv papers submitted each year from 2018 to 2025 were utilized for this part of the analysis. We also used the following two simple prompts to examine the differences between original arXiv abstracts and those revised by \texttt{GPT-4o-mini} (\(\text{temperature}=1\), \(\text{top-p}=0.9\)):
\begin{itemize}
    \item (\textbf{P1}) \textit{Revise the following sentences: \dots}
    \item (\textbf{P2}) \textit{Don't use the following words in your responses: 'realm', 'pivotal', 'intricate', 'showcasing'. Revise the following sentences: \dots}
\end{itemize}

\begin{figure}[htbp]
    \centering
    \includegraphics[width=0.8\linewidth]{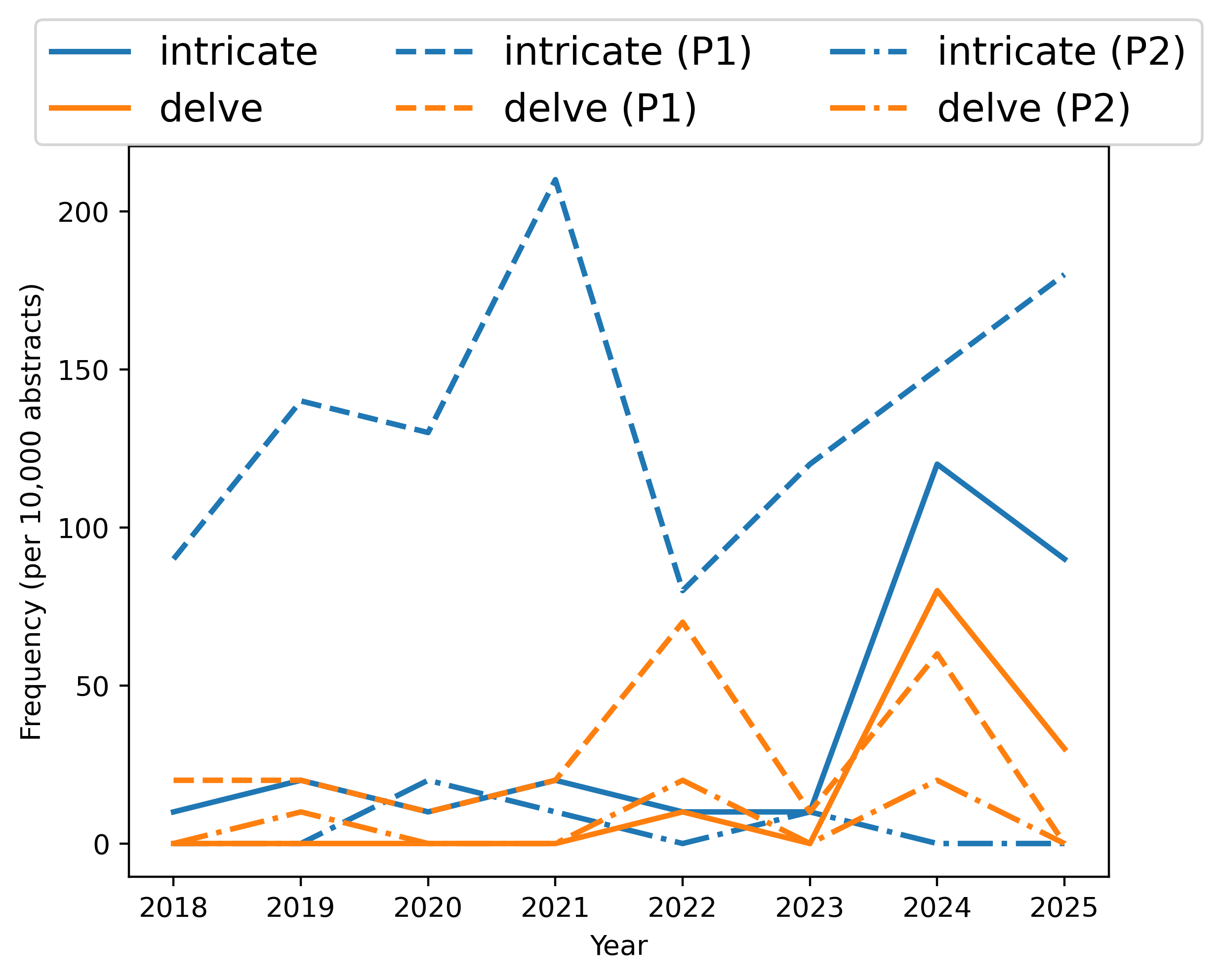}
    \includegraphics[width=0.9\linewidth]{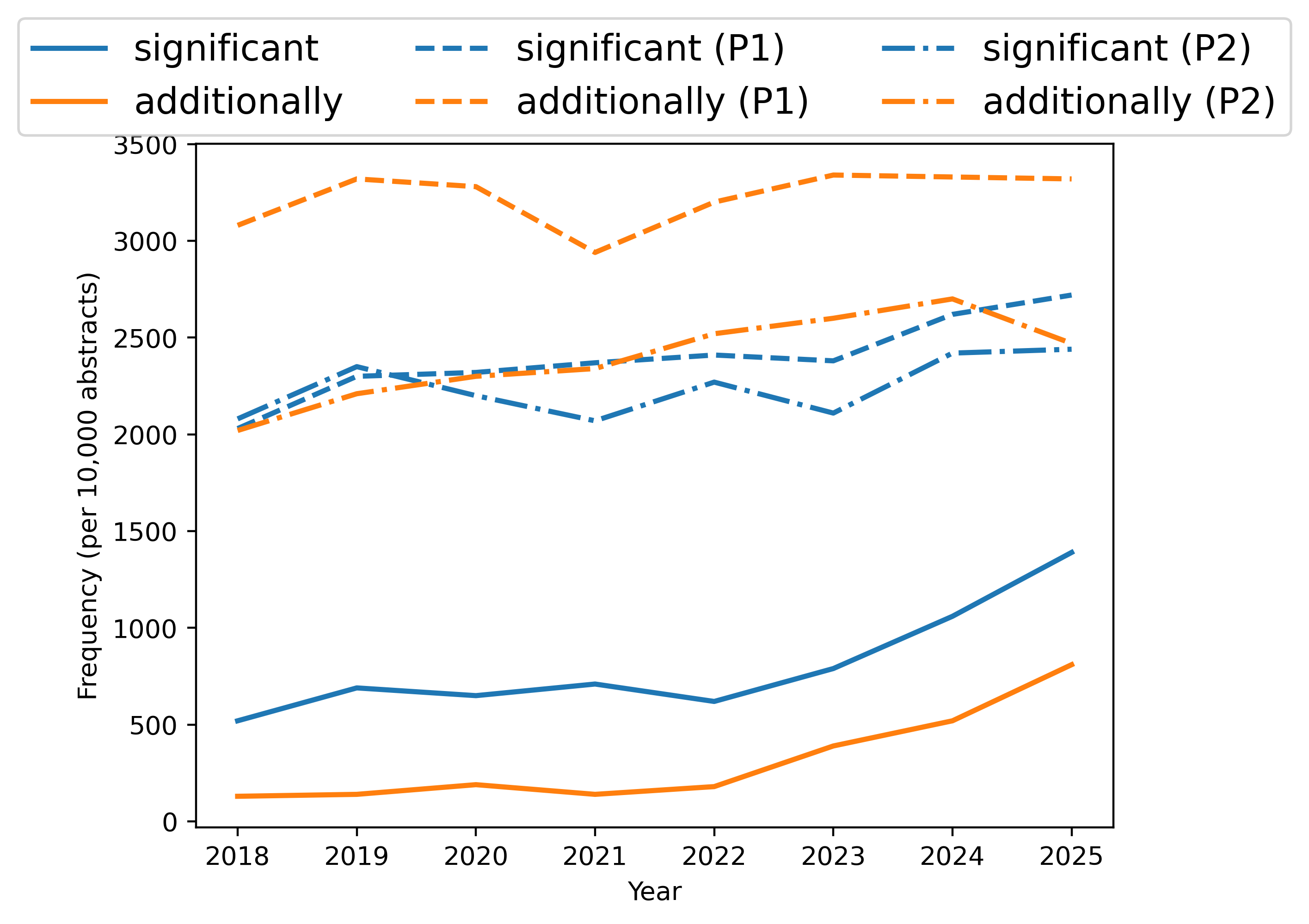}
    \caption{Comparison of word frequencies before and after LLM processing (with prompts \textbf{P1} or \textbf{P2}).}
    \vspace{-1em}
    \label{wf_detect_compare}
\end{figure}

The results in Figure~\ref{wf_detect_compare} reinforce the point that the frequency of certain words increases after LLMs revision. Using prompt \textbf{P2}, aimed at suppressing them, reduces the frequency of such words, although it does not completely eliminate them.

\begin{figure}[htbp]
    \centering
    \begin{subfigure}[b]{0.45\textwidth}
        \centering
        \includegraphics[width=\textwidth]{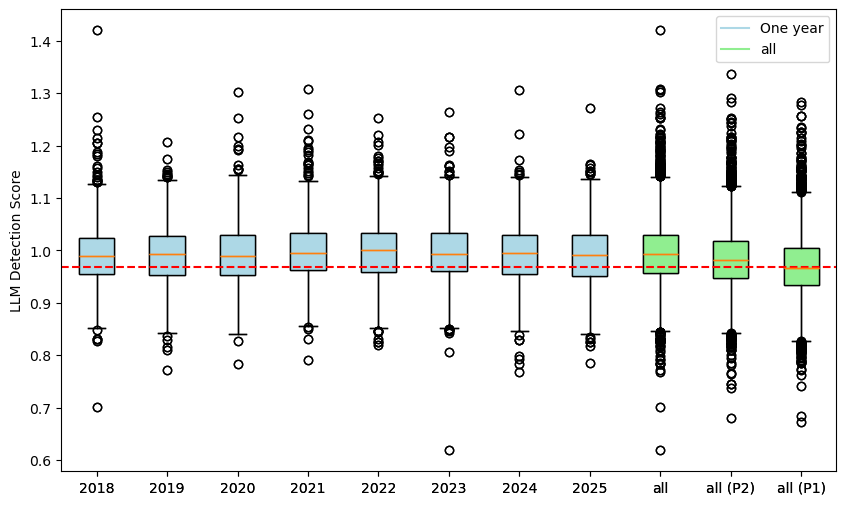}
        \caption{The last 3 columns all include abstracts of 8 years.}
    \end{subfigure}
    \hfill
    \centering
    \begin{subfigure}[b]{0.45\textwidth}
        \centering
        \includegraphics[width=\textwidth]{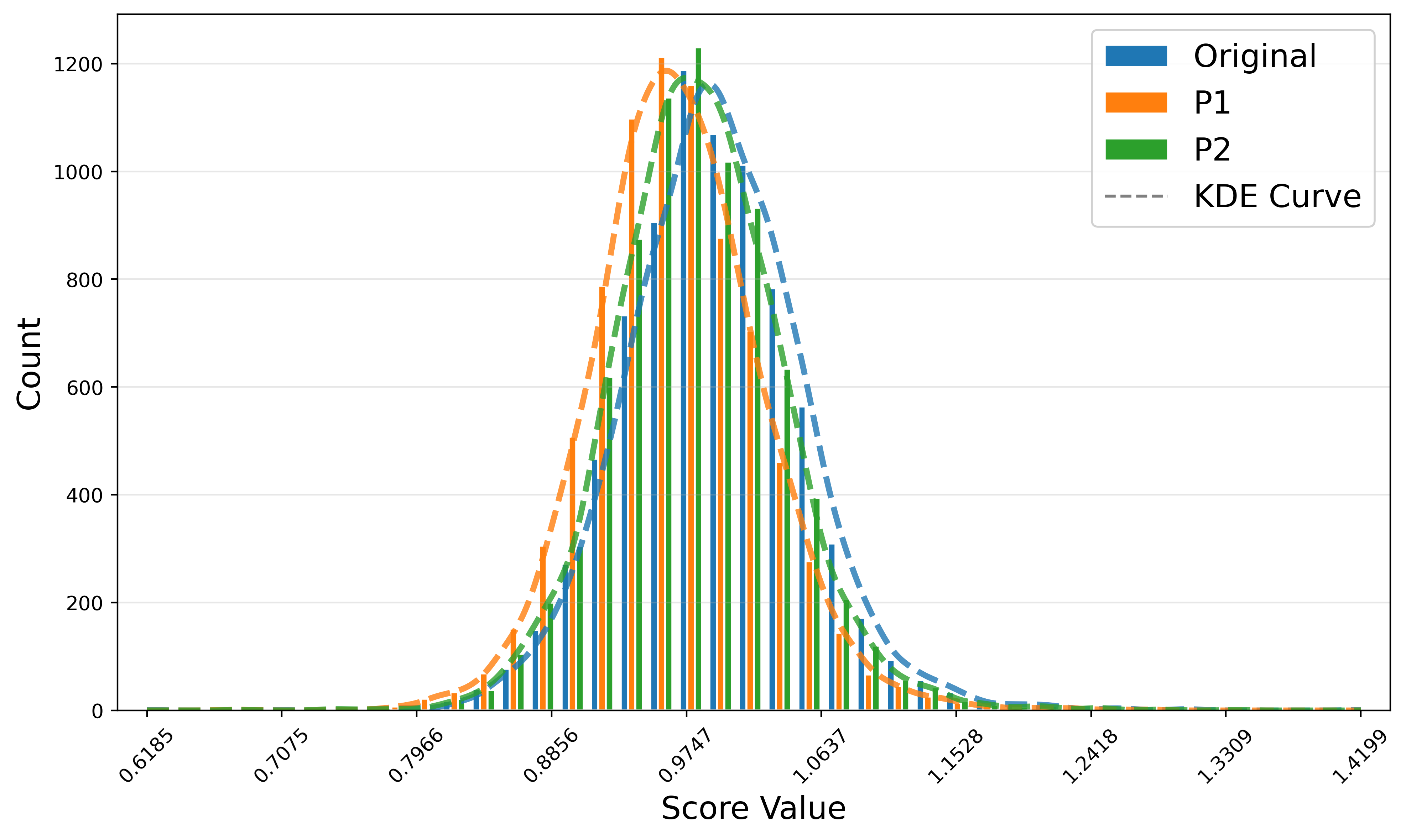}
        \caption{KDE means kernel density estimation.}
    \end{subfigure}
    \caption{MGT detection results for real and LLM-processing abstracts (with prompts \textbf{P1} or \textbf{P2}). A lower score indicates a greater probability that the text is machine-generated.}
    \label{bino_compare}
    \vspace{-1em}
\end{figure}

Figure~\ref{bino_compare} presents the detection results based on Binoculars~\citep{hans2024spotting}, one of the state-of-art MGT detectors, where a lower score indicates a greater probability that the text is machine-generated. Unlike the results obtained with our frequency analysis, Binoculars on average does not return any difference in score for the real abstracts of papers as a function of time. Moreover, the change in the detection score between the original abstracts and the texts processed by LLMs (true positives) is not significant. Furthermore, the prompts used for processing can influence the results of MGT detectors. These results raise doubts about the accuracy of the detectors, given that they are analyzing texts that have been fully processed by LLMs.

\section{Conclusion and Discussion}
Humans and LLMs are coevolving and we can already conclude that, for this reason,  the impact of LLMs on academic writing will fully assert itself over the long term. According to recent studies, people who frequently use ChatGPT for writing tasks can accurately distinguish AI-generated text~\citep{russell2025people}, which implies that they are also able to foil MGT detectors.

\textit{Grammarly} can sometimes achieve effects similar to those of ChatGPT~\citep{rudnicka2023can}, and the mix of human-written text and machine-generated text should be very common in academic writing. Detecting LLM-generated content with accuracy is becoming more difficult, perhaps impossible on a text-by-text basis.

Our findings suggest that some researchers may intentionally avoid using LLM-characteristic terms, but they are not as sensitive to some relatively common words. The gradual decrease in the occurrence of ``is'' and ``are'' in arXiv abstracts is an excellent example of such a trend, which we ascribe to a more subtle -- and continually increasing-- LLM influence. 

Therefore, using the frequency of more common words to measure the impact of LLMs on a vast number of publications will be more reliable, although this approach is less suitable for the precise detection of short texts.

\bibliography{custom}
\newpage

\appendix
\section{Appendix}
\label{sec:appendix}

\begin{figure}[htbp]
    \centering
    \begin{subfigure}[b]{0.45\textwidth}
        \centering
        \includegraphics[width=\textwidth]{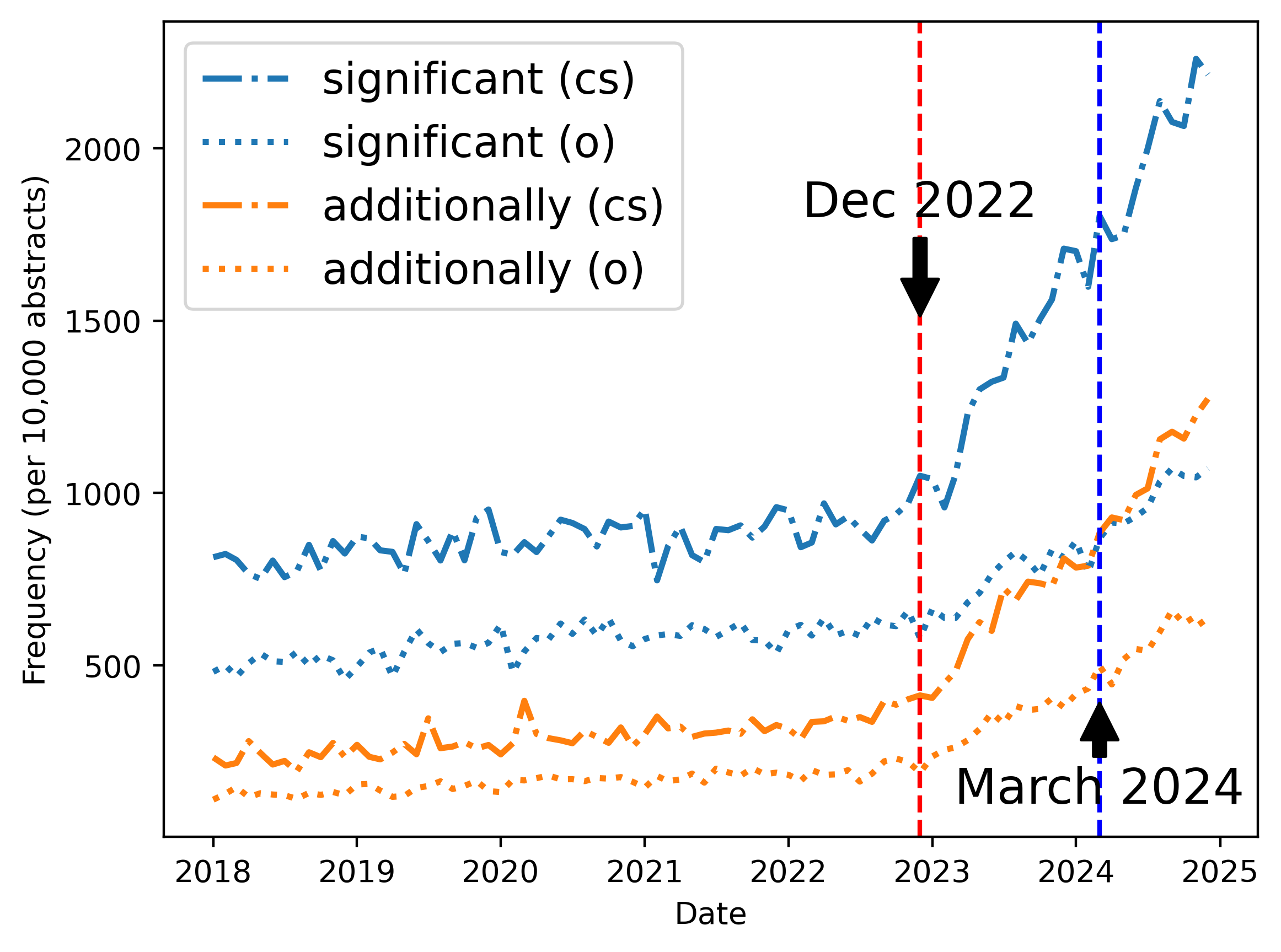}
        \caption{}
        \label{wf_significant_cs}
    \end{subfigure}
    \hfill
    \begin{subfigure}[b]{0.45\textwidth}
        \centering
        \includegraphics[width=\textwidth]{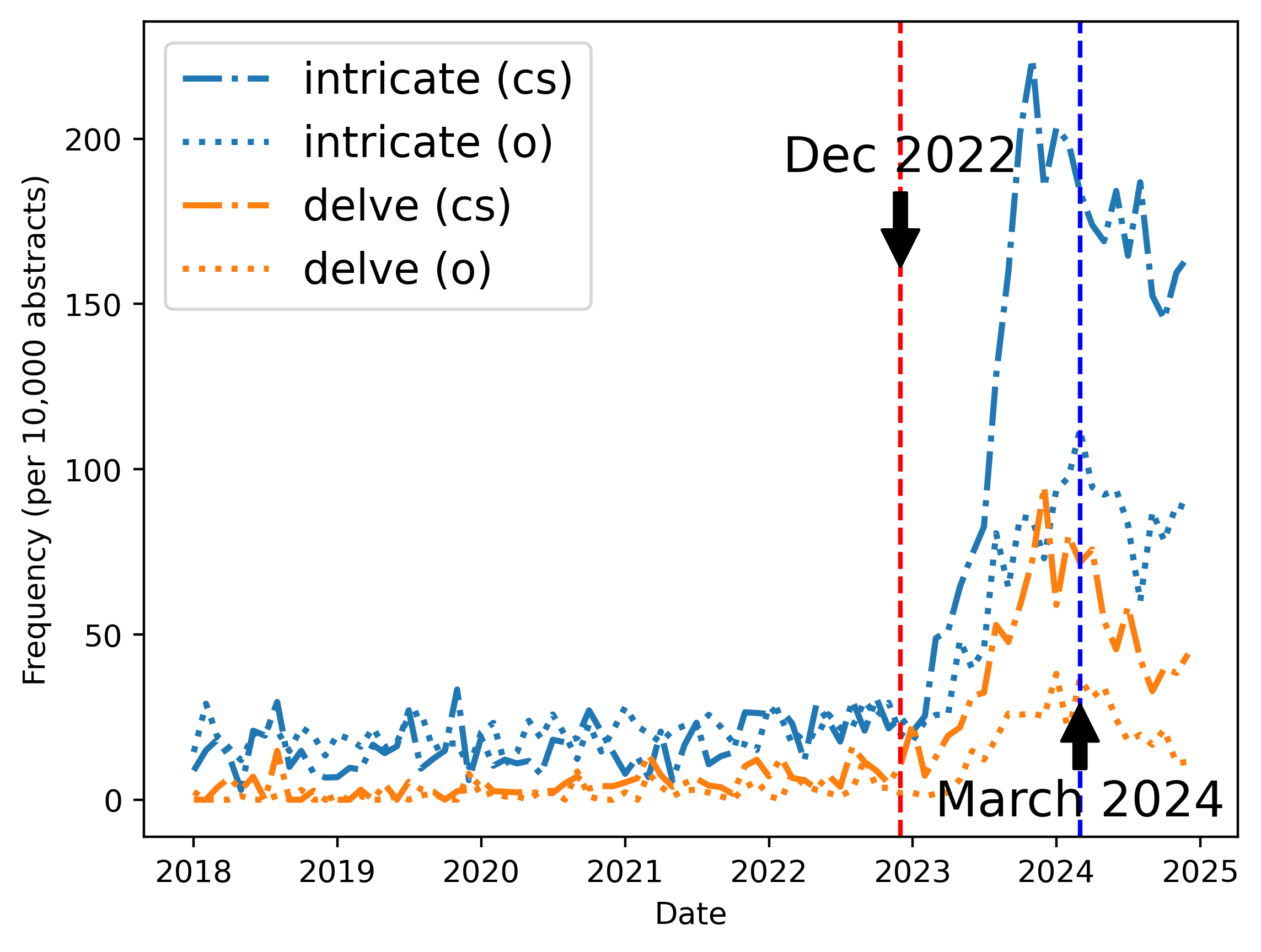}
        \caption{}
        \label{wf_intricate_cs}
    \end{subfigure}
    \hfill
    \begin{subfigure}[b]{0.45\textwidth}
        \centering
        \includegraphics[width=\textwidth]{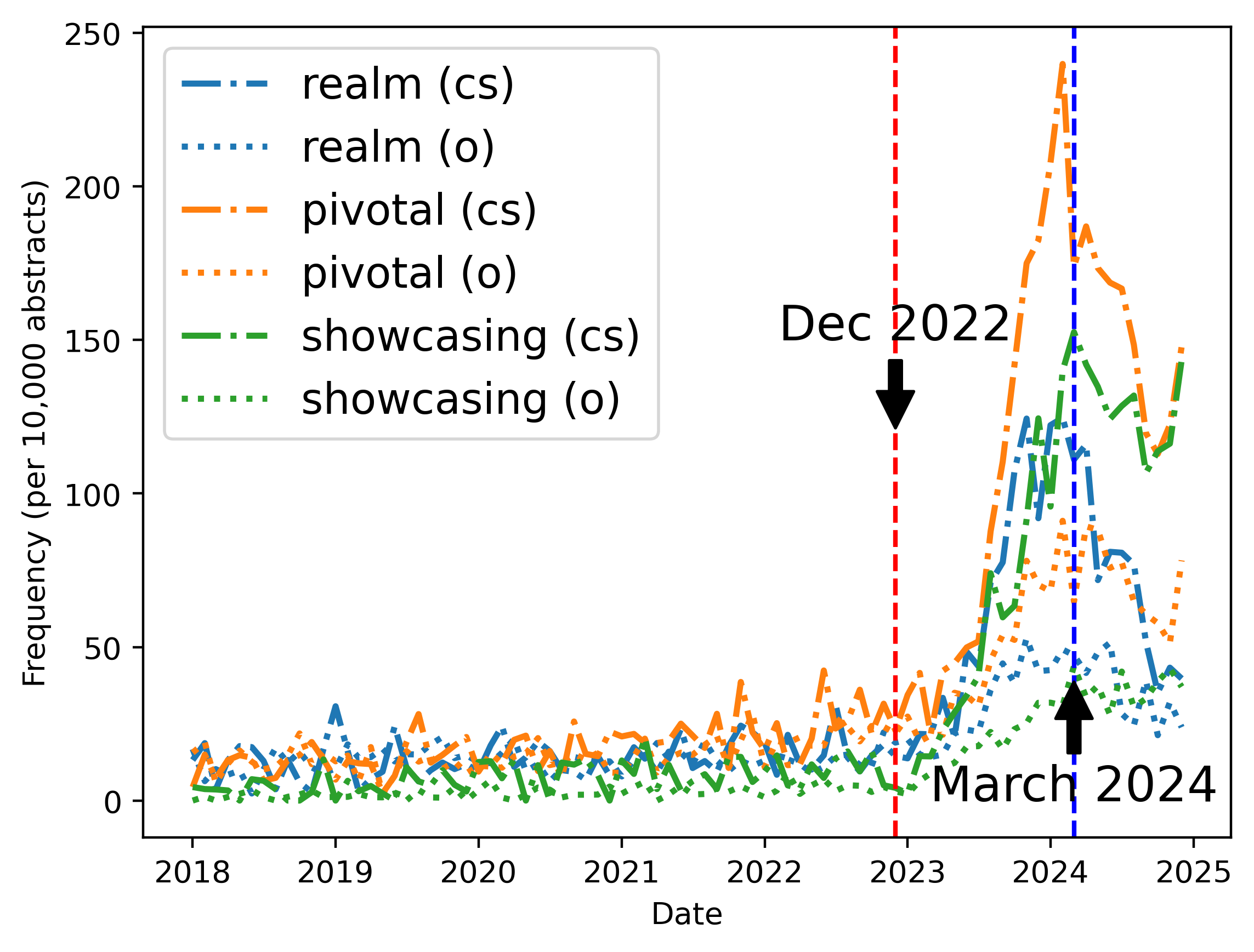}
        \caption{}
        \label{wf_realm_cs}
    \end{subfigure}
    \hfill
    \caption{Frequency of some words in arXiv abstracts.}
    \label{cs_o}
\end{figure}

\begin{figure*}[htbp]
    \centering
    \begin{subfigure}[b]{0.45\textwidth}
        \centering
        \includegraphics[width=\textwidth]{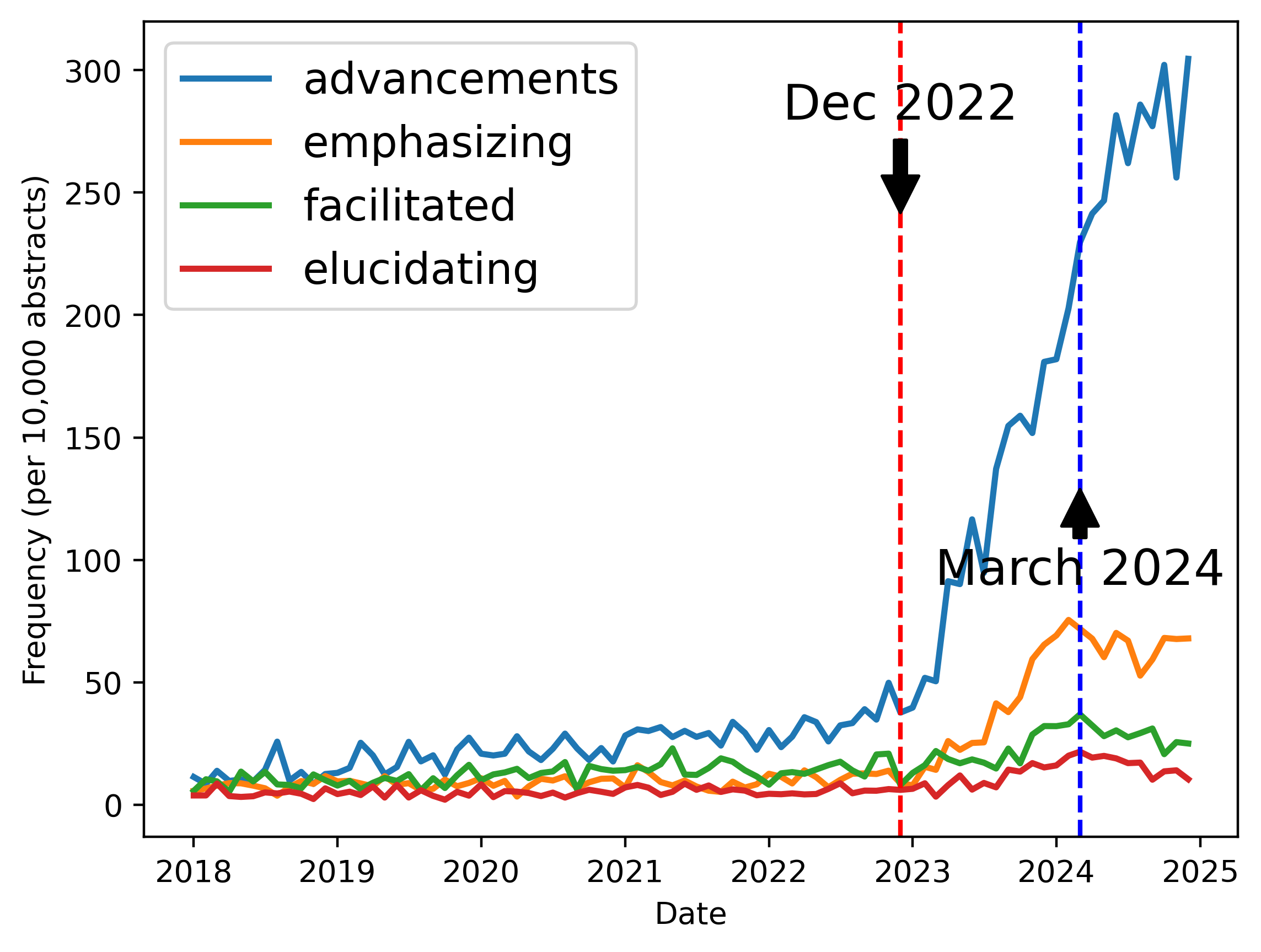}
        \caption{}
        \label{wf_advancements}
    \end{subfigure}
    \hfill
    \begin{subfigure}[b]{0.45\textwidth}
        \centering
        \includegraphics[width=\textwidth]{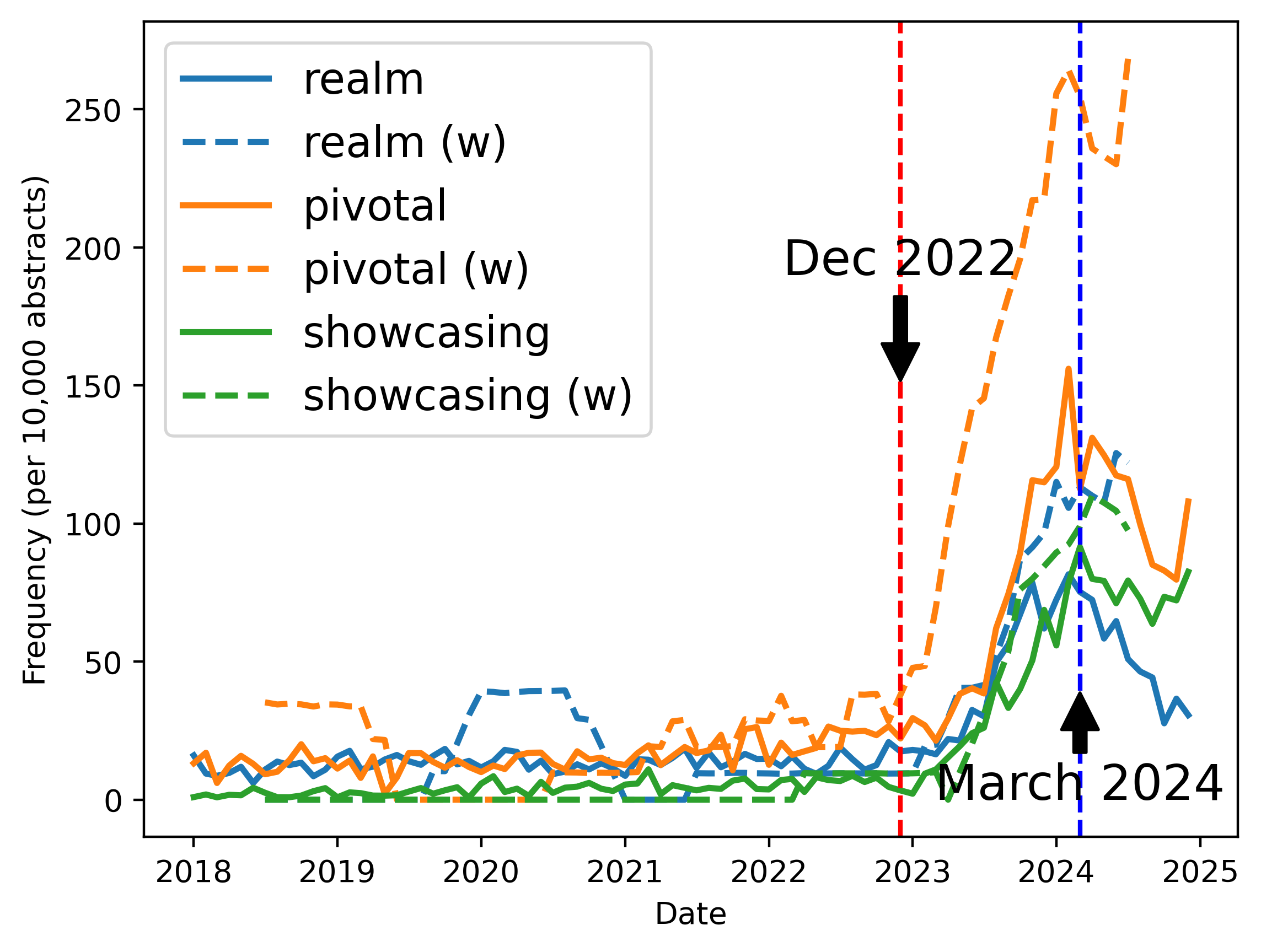}
        \caption{}
        \label{wf_realm_w}
    \end{subfigure}
    \hfill
    \begin{subfigure}[b]{0.45\textwidth}
        \centering
        \includegraphics[width=\textwidth]{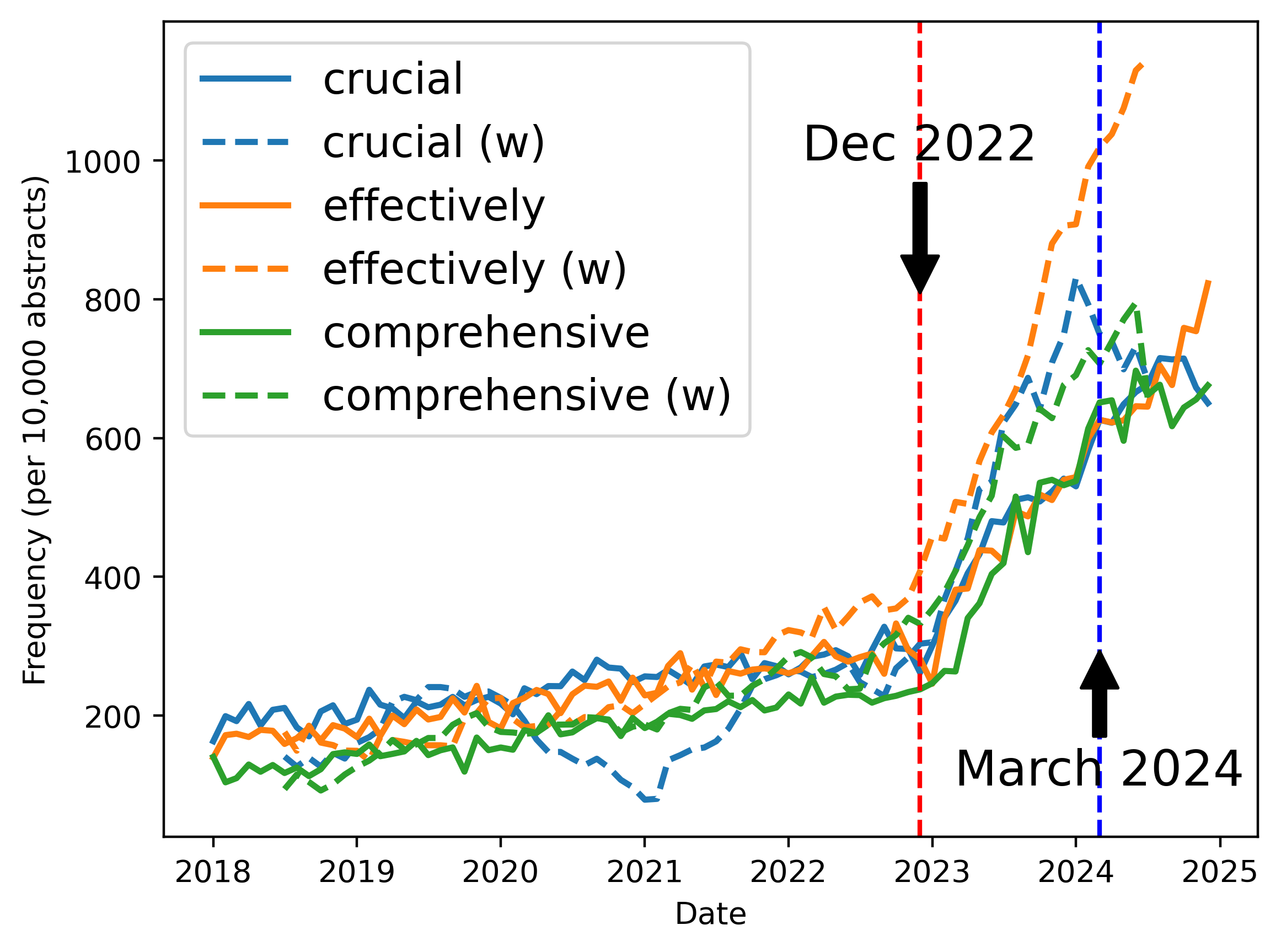}
        \caption{}
        \label{wf_crucial_w}
    \end{subfigure}
    \hfill
    \begin{subfigure}[b]{0.45\textwidth}
        \centering
        \includegraphics[width=\textwidth]{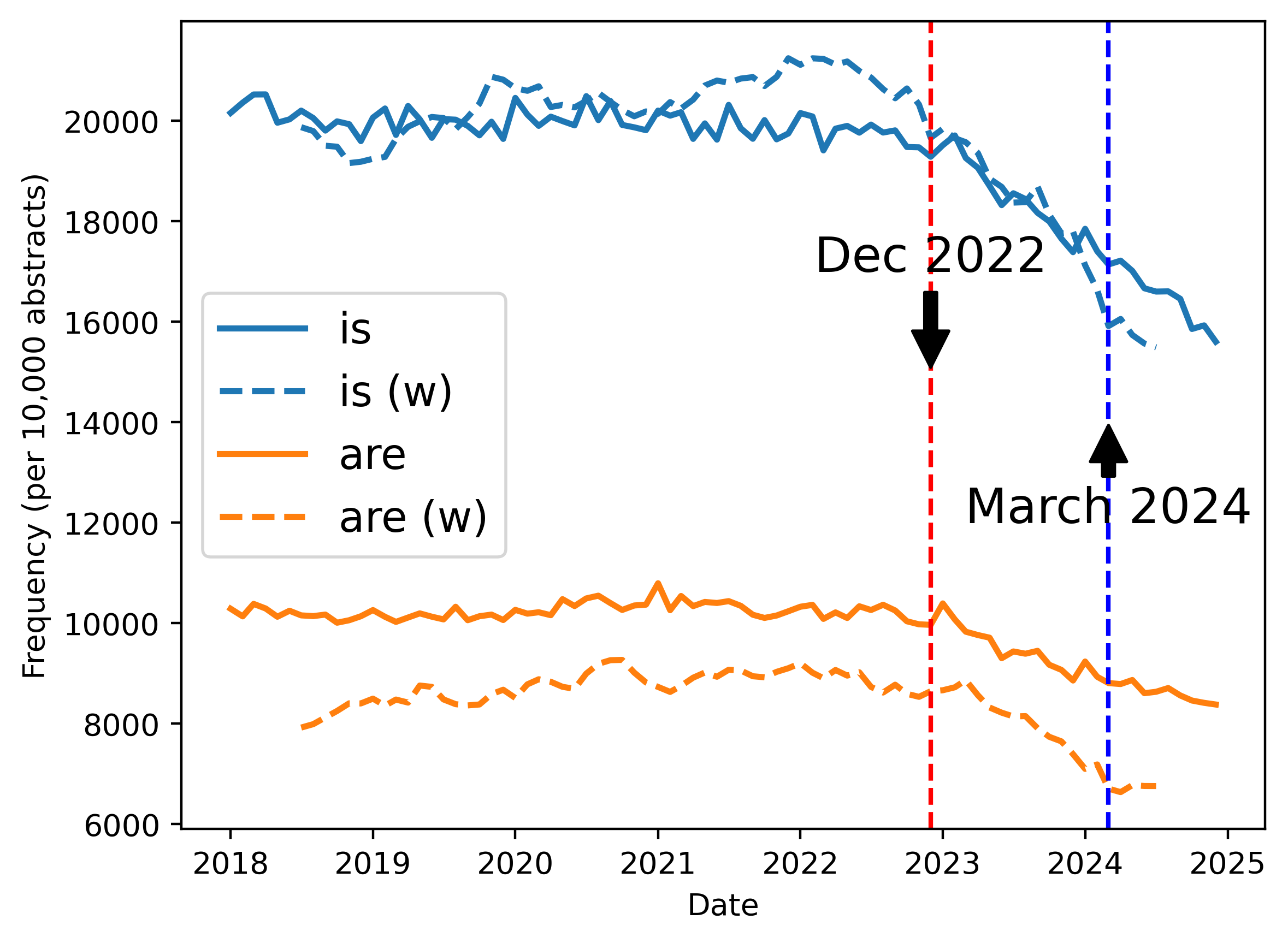}
        \caption{}
        \label{wf_is_w}
    \end{subfigure}
    \hfill
    \caption{Frequency of some words in arXiv abstracts.}
\end{figure*}

\begin{table*}
  \centering
  \begin{tabularx}{\textwidth}{llX}
    \hline
    \textbf{Paper}           & \textbf{Citations} & \textbf{Highlighted words} \\
    \hline
    \citep{liang2024monitoring}  &   87        &   \textbf{commendable, innovative, meticulous, intricate, notable, versatile.}     \\
    \citep{liang2024mapping}     &   58        &    \textbf{pivotal, intricate, realm, showcasing.}                     \\
    \cite{gray2024chatgpt}    &   41        &   words  listed based on \citet{liang2024monitoring}                    \\
    \citep{geng2024chatgpt}     &   11        & \textbf{significant, crucial,  effectively, additionally, comprehensive, enhance, capabilities, valuable.}                     \\
    \citep{liu2024towards} &   4        & \\
    \hline
  \end{tabularx}
  \caption{Papers on word frequency analysis published in March and April 2024 (submitted to arXiv). The Google citation counts are as of January 16, 2022.}
  \label{paper_compare}
\end{table*}

\end{document}